\pdfoutput=1

\documentclass[11pt]{article}

\usepackage[]{ACL2023}

\usepackage{times}
\usepackage{latexsym}
\usepackage{arydshln}

\usepackage[T1]{fontenc}

\usepackage[utf8]{inputenc}

\usepackage{microtype}

\usepackage{inconsolata}
\usepackage{graphicx}

\usepackage{CJKutf8}

\usepackage{subfig}

\title{Towards Multilingual Automatic Open-Domain Dialogue Evaluation}

\author{John Mendonça\textsuperscript{1,2,\thanks{\hspace{0.2cm}Work conducted as a visiting scholar at CMU.}},  Alon Lavie\textsuperscript{3,4} \and Isabel Trancoso\textsuperscript{1,2} \\
  \textsuperscript{1} INESC-ID, Lisbon \\
  \textsuperscript{2} Instituto Superior Técnico, University of Lisbon \\
  \textsuperscript{3} Carnegie Mellon University, Pittsburgh \\
  \textsuperscript{4} Phrase, Pittsburgh \\
  \texttt{\{john.mendonca, isabel.trancoso\}@inesc-id.pt} \\
  \texttt{alavie@cs.cmu.edu} \\}

\begin{document}
\maketitle
\begin{abstract}

The main limiting factor in the development of robust multilingual open-domain dialogue evaluation metrics is the lack of multilingual data and the limited availability of open-sourced multilingual dialogue systems. In this work, we propose a workaround for this lack of data by leveraging a strong multilingual pretrained encoder-based Language Model and augmenting existing English dialogue data using Machine Translation. We empirically show that the naive approach of finetuning a pretrained multilingual encoder model with translated data is insufficient to outperform the strong baseline of finetuning a multilingual model with only source data. Instead, the best approach consists in the careful curation of translated data using MT Quality Estimation metrics, excluding low quality translations that hinder its performance.

\end{abstract}

\section{Introduction}
\label{sec:intro}

Open-domain dialogue systems have gained substantial attention in the NLP (Natural Language Processing) and ML (Machine Learning) fields, thanks to their increasingly human-like behaviour \citep{thoppilan2022lamda,Shuster2022BlenderBot3A}. Their impressive generation capabilities can be attributed to new milestones in model development and scaling \citep{adiwardana2020towards}, and the amount of data used during training. Despite this research and development effort, advertised generation capabilities were only attainable in a select few languages (typically English or Chinese) due to low resources in dialogue for other languages \citep{mdia}. More recently, however, the advent of LLMs (Large Language Models) finetuned with Reinforcement Learning from Human Feedback such as ChatGPT \cite{ouyang2022training} has opened the path for high-quality and easily accessible multilingual dialogue generation.

\begin{figure}[ht]
  \centering
  \includegraphics[width=0.48\textwidth]{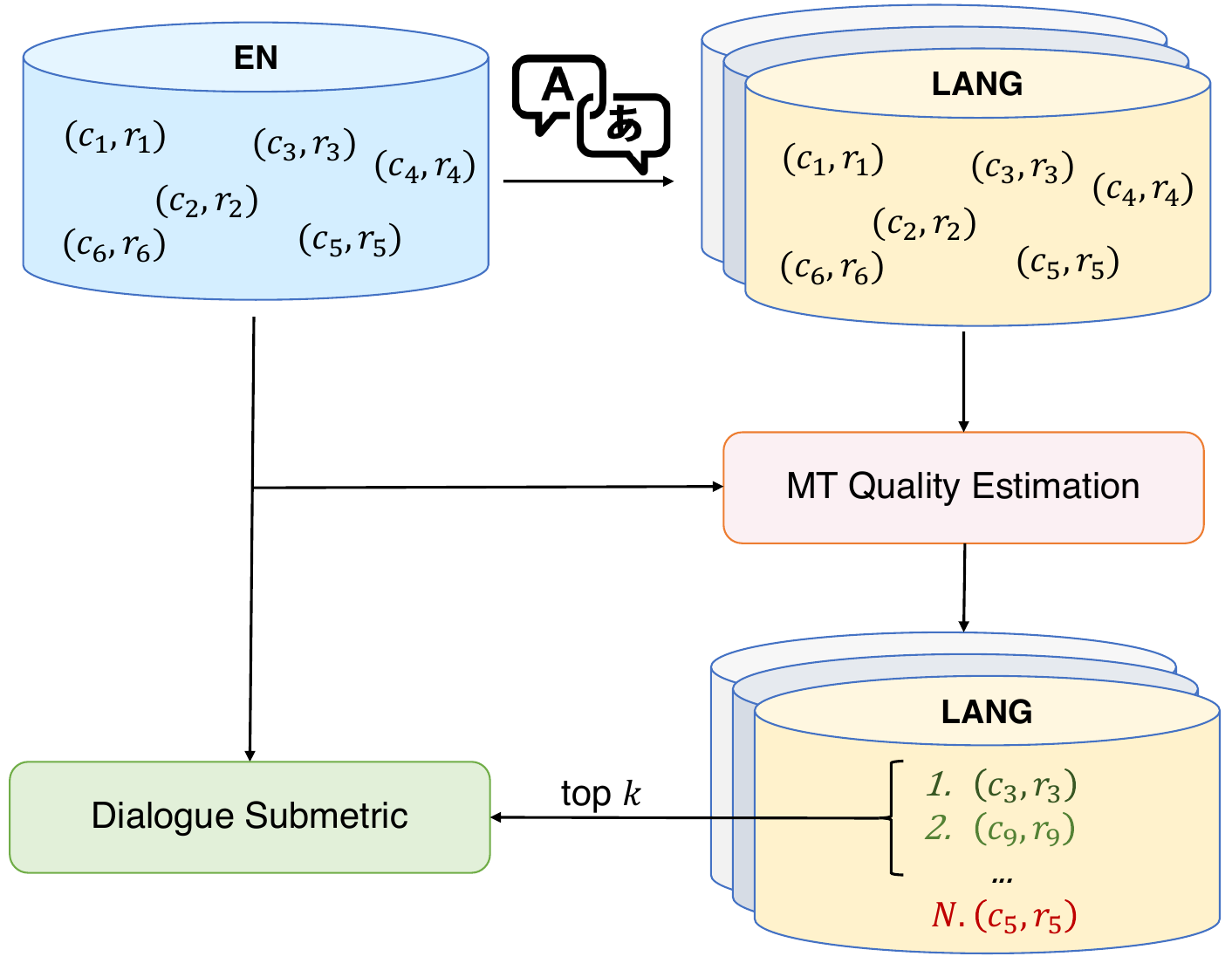}
  \caption{Proposed architecture. The original dialogue dataset is transformed into context-response pairs $(c_n,r_n)$ and translated using MT. The final dialogue submetric is trained using a combination of the original English data and the top $k$ sentences or $(c_n,r_n)$ from each language, depending on the submetric.}
  \label{fig:diagram}
\end{figure}

Similarly, automated open-domain dialogue evaluation has also been largely limited to evaluating a select few languages. Word-overlap based metrics from NLG (Natural Language Generation) such as BLEU \citep{10.3115/1073083.1073135} and METEOR \citep{banerjee-lavie-2005-meteor} are agnostic to language, only requiring a reference response. However, these metrics are known to correlate poorly with human judgments due to the multifaceted nature of dialogue \citep{liu-etal-2016-evaluate}. Reference-free metrics such as USR \citep{mehri-eskenazi-2020-usr} and USL-H \citep{phy-etal-2020-deconstruct}, however, require dialogue data for training. Considering most open-source dialogue data is in English, these models are expected to underperform significantly in other languages. Additionally, most open sourced dialogue systems are also limited to English, further disincentivising multilingual research.

One solution to the issues previously mentioned is to leverage MT (Machine Translation). With MT services becoming more affordable and consistent, some authors resort to translation when developing their multilingual dialogue systems \citep{schuster-etal-2019-cross-lingual,anastasiou-etal-2022-machine}.
This can either be included as a module in the system's pipeline -- allowing the use of proven English generation models for other languages; or as a cross-lingual transfer method -- by translating training data.

In this paper, we extend the approach of training using data generated by MT for the development of multilingual models for evaluation of open-domain dialogue responses. 
We experiment with and evaluate several different possible workarounds for this problem. Namely, we leverage the availability of strong pretrained multilingual encoders as a foundation for training multilingual dialogue evaluation models.
As a first step, we translate existing publicly-available English dialogue data into the target languages. We then explore multiple alternative ways to leverage this translated data in order to finetune and train monolingual and multilingual dialogue evaluation models for two specific dialogue submetrics. To address the impact of low quality translations, we propose using an MT Quality Estimation (QE) model to rank the translations and investigate the impact of finetuning models with varying amounts of quality-ranked data. Figure \ref{fig:diagram} illustrates the proposed approach. 

The performance of these alternative models is evaluated on a curated test set of dialogues which were human-annotated with dialogue quality scores for two subqualities. The original English test set was translated using MT and then post-edited by editors into six different target languages (PT-Portuguese, DE-German, FR-French, ZH-Chinese, ES-Spanish and JA-Japanese). The quality scores from the human annotations of the original English dialogues were then carried over to the target-language dialogues. Our finetuned multilingual dialogue evaluation models exhibit strong correlations with human judgements, comparable to LLMs, indicating it is possible to leverage multilingual dialogue evaluation metrics without the constraints LLMs currently possess (costs, latency, etc.). We hope this will encourage other researchers to update existing metrics using our proposed multilingual finetuning approach.

In summary, the primary contributions of this work are as follow:

\begin{itemize}
    \item We evaluate cross-lingual transfer and translation augmented training approaches using MT for the task of training multilingual dialogue evaluation models, showing that, on average, the best performance is achieved by finetuning with subsets consisting of only the best translations. We found that, depending on the subquality and target language, the optimal amount of translated data can be as low as 5\% and as high as 75\%. 
    \item We translate and release DailyDialog and a corresponding test set of human quality annotations in 6 languages to facilitate future benchmarking of multilingual dialogue evaluation metrics\footnote{\url{github.com/johndmendonca/DialEvalML}}.
\end{itemize}

\section{Background}
\label{sec:background}

\subsection{Open-Domain Dialogue Evaluation Metrics}



The recent trend in open-domain dialogue evaluation is to train dialogue submetrics using well-defined self-supervised tasks which correlate well with their corresponding subqualities. The most used self-supervised task is Next Sentence Prediction (NSP), as it is known to correlate well with subqualities that evaluate \textit{"Context Awareness"}. Examples of this include: \textit{Uses Context} \citep{mehri-eskenazi-2020-usr}, \textit{Sensibleness} \citep{phy-etal-2020-deconstruct,mendonca-etal-2022-qualityadapt} and \textit{Relevance}  \citep{zhao,zhang2022mme}. Other subqualities include: \textit{Fluency, Grammatically Correct} or \textit{Understandability}, which use word-level noising techniques to generate negative samples \citep{phy-etal-2020-deconstruct,mendonca-etal-2022-qualityadapt,zhang2022mme}; and \textit{Specificity}, which uses an MLM (Masked Language Modelling) score \citep{mehri-eskenazi-2020-usr,phy-etal-2020-deconstruct,zhang2022mme}. For overall quality, these submetrics are typically combined using different methods (e.g. empirical observation, trained Linear Regression or multilayer perceptrons).

To the best of our knowledge, there has not been any published research on cross-lingual transfer and/or development of trained multilingual metrics for open-domain dialogue evaluation.

\subsection{Multilingual Text Classification}

Despite the lack of research on multilingual dialogue evaluation, extending text classification to other languages is a well established subfield of research in NLP. The main constraint for multilingual performance parity is the lack of task-specific resources in the vast majority of written languages. Given the creation of these resources is both time consuming and expensive, most research effort has been geared towards general-purpose cross-lingual representations that are learned in an unsupervised way, therefore leveraging the unstructured data available in the wild. Large multilingual Transformer-based models (e.g mBERT, XLM-RoBERTa, and mT5) have been successfully used in a variety of classification tasks \citep{conneau-etal-2020-unsupervised, pires-etal-2019-multilingual, xue-etal-2021-mt5}. The standard approach for cross-lingual transfer is to finetune on existing domain data in a source language and perform inference in a target language. However, this approach typically lags behind models specifically trained with in-domain (both task and language) data.

As a solution to this problem, \citet{pfeiffer-etal-2020-mad} propose learning language-specific adapter modules via MLM on unlabelled target-language data followed by task-specific adapter modules by optimising a target task on labelled data in the source language. Task and language adapters are stacked, allowing cross-lingual transfer to the target language by substituting the target-language adapter at inference. 

\citet{bornea2021multilingual} propose an augmentation strategy where a corpus of multilingual silver-labelled QA pairs is generated by combining the original English training data with MT-generated data. A language adversarial training and arbitration framework bring the embeddings closer to each other, making the model language invariant.

To the best of our knowledge, there has not been any research on the utilization of MT Quality Estimation (QE) scoring as a means for identifying and demoting or excluding poorly translated data in such cross-language training scenarios.

\section{Problem Formulation}
\label{sec:formulation}

The goal of reference-free turn-level dialogue evaluation is, given a dialogue history (frequently denoted as context) $c$ of varying amount of turns, and a response $r$, to learn a scoring function that assigns a score $f(c,r) \rightarrow s$. This scoring function is compared against human judgements, which annotate the same context-response pairs. These responses are evaluated using a scaling method, for instance, a binary $(0,1)$ judgement or a $[1,5]$ scale, where the lowest value means lowest quality and highest value maximum quality. The notion of quality varies wildly depending on the annotation. In this work, we evaluate dialogue in two dimensions: 

\begin{itemize}
    \item \textbf{Understandability} An understandable response is one that can be understood without context. Such responses may contain minor typos that do not hinder the comprehension of the response.
    \item \textbf{Sensibleness} A sensible response is one that takes into account its preceding context. 
\end{itemize}

Most automatic evaluation metrics reformulate the problem as regression. Performance is then evaluated using Pearson and Spearman correlations with human annotations.

\subsection{Automatic Dialogue Evaluation Metrics}

The majority of competitive metrics for dialogue evaluation include models trained in a self-supervised way for Valid Sentence Prediction (VSP) and Next Sentence Prediction (NSP) \citep{yeh-etal-2021-comprehensive, Zhang2021AutomaticEA}. As such, the focus of this work was to evaluate multilingual dynamics for these models, which can then be employed on existing metrics.

\paragraph{VSP: Valid Sentence Prediction} In this paper, we followed the approach used by \citet{phy-etal-2020-deconstruct} and initially proposed by \citet{sinha2020learning}. A regression model was trained to differentiate between positive samples and synthetic negative samples. \textbf{Positive} samples are perturbed by randomly applying one of the following: (1) no perturbation, (2) punctuation removal, (3) stop-word removal. \textbf{Negative} samples are generated by randomly applying one of the following rules: (1) word reorder (shuffling the ordering of the words); (2) word-drop; and (3) word-repeat (randomly repeating words).

\paragraph{NSP: Next Sentence Prediction} The task of predicting sensibleness can be considered a binary (NSP) task, distinguishing a positive example from a semantically negative one, given a context. A discriminative regression model was trained using the following sampling strategy: \textbf{positive} responses are drawn directly from the dialog; \textbf{negative} responses are randomly selected and a token coverage test discards semantically similar sentences. All responses are processed using the positive-sample heuristic used by VSP.

\section{Cross-lingual Transfer Learning}
\label{sec:naive_exp}

The goal of the experiments described in this section was to evaluate different basic approaches of cross-lingual transfer for the task of automatic dialogue evaluation. For encoder model training, we leveraged Machine Translation (MT) by fully translating an English source dialogue dataset and then finetuning monolingual and multilingual models using these translations.

\subsection{Experimental Setup}

\subsubsection{Dataset}

All experiments in this paper were based on the \textbf{DailyDialog \citep{li-etal-2017-dailydialog}} dataset, a high-quality human-human open-domain dialogue dataset focused on day-to-day conversations. After processing, we obtained train/dev splits of 58,515/25,078 and 89,707/38,449 per language for the VSP and NSP models, respectively. For training and evaluation, the post-processed dataset was translated into the target languages using MBART50 \cite{liu-etal-2020-multilingual-denoising}. We opted for using MBART50 as it is a relatively lightweight open sourced model with a large language coverage.

For the test set, we leveraged the annotations from \citet{phy-etal-2020-deconstruct}. These human annotations evaluate five responses from two retrieval methods, two generative methods, and one human-generated response for 50 contexts. 
These responses were annotated in terms of \textit{Understandability} and \textit{Sensibleness} \footnote{Annotations for \textit{Specificity} and \textit{Overall Quality} were also conducted, but were excluded since they do not map to the learned metrics under study.}. We translated this set using Unbabel's\footnote{unbabel.com} translation service. A total of 300 sentences were translated, corresponding to the 50 shared contexts and 250 responses. 
The translations were then split into smaller tasks and were corrected by editors from a commercial provider. Editors were specifically asked to retain any source disfluencies or hallucinations stemming from low quality response generation (e.g. \textit{"I'm afraid you can't. I'm afraid you can't."}; \textit{"Au contraire, you need to be a bahn."}). This ensured the original human quality annotations remained valid for the translation. A secondary senior editor reviewed the edited content as a whole.

\subsubsection{Finetuned Encoders}

We used XLM-RoBERTa \cite{conneau-etal-2020-unsupervised} as the encoder model for the experiments. This model is the multilingual version of RoBERTa, pretrained on CommonCrawl data containing 100 languages. For both the VSP and NSP models, we added a regression head on top of the encoder model.

\paragraph{EN -- Zero-shot inference} As a baseline for our results, we conducted zero-shot inference on the target languages using a model finetuned only on the original English data.

\paragraph{LANG -- Target-Language Finetuning}

We finetuned the encoder with target-language translated dialogue data only. The downside of this approach is that a unique model needs to be trained for each target language. However, this method can be scaled to every language, including new ones, and is optimised to perform best in that language. 

\begin{table*}[ht]
\centering
\scriptsize
\begin{tabular}{lcccccccccccccccc}
\multicolumn{1}{l|}{}               & \multicolumn{2}{c|}{\textbf{EN}}                   & \multicolumn{2}{c|}{\textbf{PT}}                   & \multicolumn{2}{c|}{\textbf{DE}}                   & \multicolumn{2}{c}{\textbf{FR}}                    & \multicolumn{2}{c|}{\textbf{ZH}}                   & \multicolumn{2}{c|}{\textbf{ES}}                   & \multicolumn{2}{c|}{\textbf{JA}}                   & \multicolumn{2}{c}{\textbf{AVG}} \\
\multicolumn{1}{l|}{}               & \textbf{Pr.}  & \multicolumn{1}{c|}{\textbf{Sp.}}  & \textbf{Pr.}  & \multicolumn{1}{c|}{\textbf{Sp.}}  & \textbf{Pr.}  & \multicolumn{1}{c|}{\textbf{Sp.}}  & \textbf{Pr.}  & \multicolumn{1}{c|}{\textbf{Sp.}}  & \textbf{Pr.}  & \multicolumn{1}{c|}{\textbf{Sp.}}  & \textbf{Pr.}  & \multicolumn{1}{c|}{\textbf{Sp.}}  & \textbf{Pr.}  & \multicolumn{1}{c|}{\textbf{Sp.}}  & \textbf{Pr.}    & \textbf{Sp.}   \\ \hline
\multicolumn{17}{c}{\textbf{Understandability}}                                                                                                                                                                                                                                                                                                                                                                                                           \\ \hline
\multicolumn{1}{l|}{\textbf{EN}}    & .376 & \multicolumn{1}{c|}{.187} & \textbf{.366} & \multicolumn{1}{c|}{.167} & .328 & \multicolumn{1}{c|}{.172} & .351 & \multicolumn{1}{c|}{.120} & \textbf{.318} & \multicolumn{1}{c|}{\textbf{.202}} & .342 & \multicolumn{1}{c|}{.204} &\textbf{ \textbf{.204}} & \multicolumn{1}{c|}{.176} & .327   & .194  \\
\multicolumn{1}{l|}{\textbf{LANG}}  & -             & \multicolumn{1}{c|}{-}             & .176          & \multicolumn{1}{c|}{.164}          & .214          & \multicolumn{1}{c|}{.138}          & \textit{.052} & \multicolumn{1}{c|}{.034}          & .274          & \multicolumn{1}{c|}{.156}          & .219          & \multicolumn{1}{c|}{.144}          & .185          & \multicolumn{1}{c|}{.132}          & .214            & .146           \\
\multicolumn{1}{l|}{\textbf{ML}}    & .336          & \multicolumn{1}{c|}{.117}          & .176          & \multicolumn{1}{c|}{.167}          & .262          & \multicolumn{1}{c|}{.150}          & \textit{.012} & \multicolumn{1}{c|}{.015}          & .225          & \multicolumn{1}{c|}{.138}          & .117          & \multicolumn{1}{c|}{.158}          & \textit{.091} & \multicolumn{1}{c|}{\textit{.092}} & .174            & .126           \\
\multicolumn{1}{l|}{\textbf{MAD-X}} & .363          & \multicolumn{1}{c|}{.166}          & .189          & \multicolumn{1}{c|}{.103}          & .237          & \multicolumn{1}{c|}{.122}          & .168          & \multicolumn{1}{c|}{\textit{.078}} & .305          & \multicolumn{1}{c|}{.168}          & .217          & \multicolumn{1}{c|}{.119}          & .119          & \multicolumn{1}{c|}{.129}          & .228            & .126           \\  \hdashline
\multicolumn{1}{l|}{\textbf{ChatGPT}} & \textbf{.397} & \multicolumn{1}{c|}{\textbf{.334}}          & .365          & \multicolumn{1}{c|}{\textbf{.230}}          & \textbf{.332}          & \multicolumn{1}{c|}{\textbf{.263}}          & \textbf{.369}          & \multicolumn{1}{c|}{\textbf{.273}}          & .276          & \multicolumn{1}{c|}{.182}          & \textbf{.394} & \multicolumn{1}{c|}{\textbf{.263}}          & \textbf{.228}          & \multicolumn{1}{c|}{\textbf{.223}}          & \textbf{.337}            & \textbf{.263 }          \\ \hline
\multicolumn{17}{c}{\textbf{Sensibleness}}                                                                                                                                                                                                                                                                                                                                                                                                                \\ \hline
\multicolumn{1}{l|}{\textbf{EN}}    & .658          & \multicolumn{1}{c|}{.676}          & .636          & \multicolumn{1}{c|}{.651}          & .657          & \multicolumn{1}{c|}{.655}          & .646 & \multicolumn{1}{c|}{.656}          & .640          & \multicolumn{1}{c|}{.656}          & .646          & \multicolumn{1}{c|}{.657}          & .590          & \multicolumn{1}{c|}{.599}          & .639            & .649           \\
\multicolumn{1}{l|}{\textbf{LANG}}  & -             & \multicolumn{1}{c|}{-}             & \textbf{.649} & \multicolumn{1}{c|}{.661}          & .669 & \multicolumn{1}{c|}{\textbf{.699}} & .635          & \multicolumn{1}{c|}{.655}          & .634          & \multicolumn{1}{c|}{\textbf{.671}} & .629          & \multicolumn{1}{c|}{.669} & .617 & \multicolumn{1}{c|}{\textbf{.640}} & .642   & \textbf{.664}  \\
\multicolumn{1}{l|}{\textbf{ML}}    & .651          & \multicolumn{1}{c|}{.691} & .606          & \multicolumn{1}{c|}{\textbf{.675}} & .634          & \multicolumn{1}{c|}{.680}          & .605          & \multicolumn{1}{c|}{\textbf{.669}} & .642 & \multicolumn{1}{c|}{.667}          & .596          & \multicolumn{1}{c|}{.676}          & .599          & \multicolumn{1}{c|}{.637}          & .619            & \textbf{.664}           \\
\multicolumn{1}{l|}{\textbf{MAD-X}} & .660 & \multicolumn{1}{c|}{.681}          & .614          & \multicolumn{1}{c|}{.604}          & .664          & \multicolumn{1}{c|}{.652}          & .624          & \multicolumn{1}{c|}{.624}          & .608          & \multicolumn{1}{c|}{.647}          & \textbf{.688} & \multicolumn{1}{c|}{.661}          & .558          & \multicolumn{1}{c|}{.595}          & .631            & .638           \\  \hdashline
\multicolumn{1}{l|}{\textbf{ChatGPT}} & \textbf{.746} & \multicolumn{1}{c|}{\textbf{.724}}          & .636          & \multicolumn{1}{c|}{.626}          & \textbf{.683}          & \multicolumn{1}{c|}{.675}          & \textbf{.695}          & \multicolumn{1}{c|}{.666}          & \textbf{.655}          & \multicolumn{1}{c|}{.645}          & .680 & \multicolumn{1}{c|}{\textbf{.677}}          & \textbf{.625}          & \multicolumn{1}{c|}{.610}          & \textbf{.674}            & .662       \\ \hline
\end{tabular}
\caption{Average correlation results across 3 runs with different seeds. \textbf{Pr.} denotes Pearson and \textbf{Sp.} denotes Spearman. \textbf{Bold} denotes best performance, \textit{Italic} $p < 0.05$.}
\label{tab:main-res}
\end{table*}

\paragraph{ML -- Multilingual Finetuning}

Instead of finetuning a new model for each target language, one can finetune a single multilingual model by combining all of the translated data. In this case, the resulting single trained model is then used to evaluate responses in all languages. However, its performance may suffer in languages it has not seen during finetuning, even if they are supported by the encoder model. Furthermore, unlike target-language finetuned, the multilingual model is optimised jointly for all included languages.

\paragraph{MAD-X}

In this approach, we trained a VSP and NSP task adapter using the original English data by stacking the task adapter with a pretrained English language adapter (kept frozen during training). For zero-shot inference, the English language adapter was replaced by the target-language counterpart, while keeping the trained task adapter in place.

\subsubsection{Large Language Model}

As an additional strong baseline, we leveraged \texttt{gpt-3.5-turbo} (colloquially known as ChatGPT) as an evaluator of Understandability and Sensibleness. The context (exclusively for Sensibleness) and response was provided as input, together with the prompt \textit{"\{Given the context,\} evaluate from 1-5 the response in terms of \{dimension\}. Provide the score and nothing else."}. This prompt, paired with a temperature setting of 0.0 attempted to minimises the variability of the output. Nevertheless, we report a standard deviation of (.003, .003) and (.001, .001) for Understandability and Sensibleness correlations, respectively, across 3 runs.

\subsection{Results}

The correlation results for all subqualities and the overall quality are presented in Table \ref{tab:main-res}. 

\paragraph{Understandability} The results show that, on average, the best performing encoder approach is the zero-shot inference using the English model (\textbf{EN}). Both the target-language finetuning (\textbf{LANG}) and multilingual finetuning approaches (\textbf{ML}) have much lower performances, indicating that translation augmentation is detrimental for this task. We also note that the \textbf{MAD-X} approach, although performing slightly better than ML and LANG, still lags behind EN considerably. In any case, ChatGPT largely outperforms other models on both metrics.

\paragraph{Sensibleness} The best performing encoder approach for this subquality is LANG. Intuitively this makes sense, given that during finetuning the model is exposed to target-language data for the language it is being evaluated on. Furthermore, the performance difference between the different approaches is relatively much smaller, which indicates the Sensibleness subquality is less sensitive to MT quality. When comparing these results with ChatGPT, we observe a much smaller performance gap, with the best encoder models slightly outperforming on Spearman.

\begin{figure*}
  \centering
  \begin{tabular}[b]{c}
    \includegraphics[width=.44\linewidth]{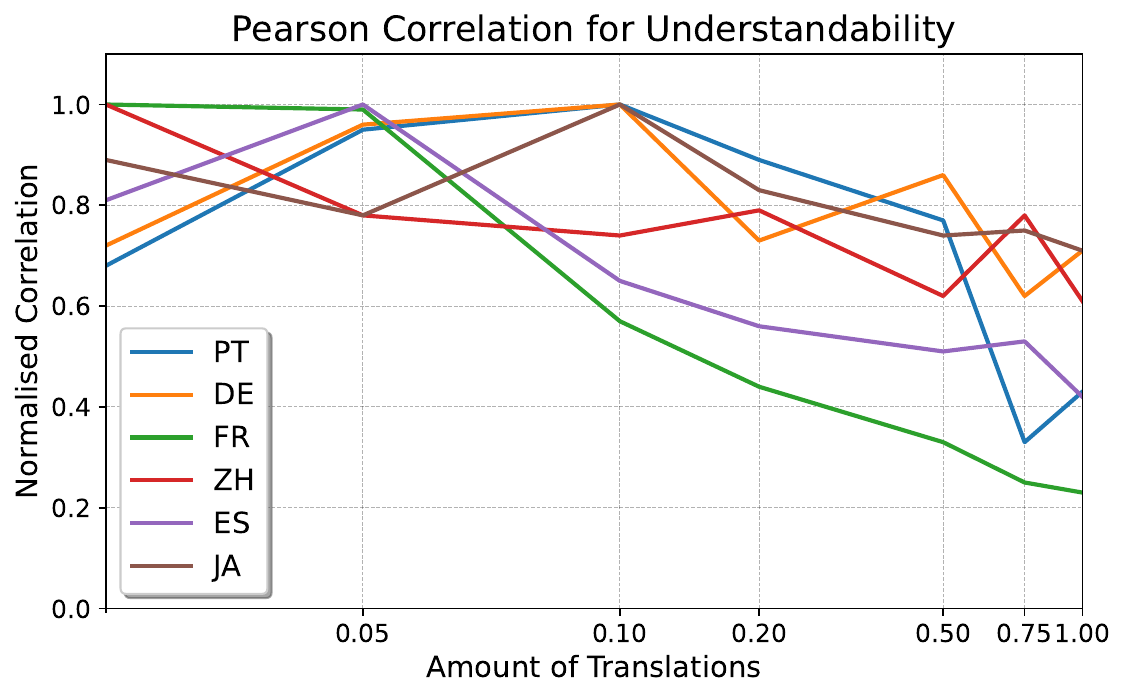} \\
    \small (a) Pearson Correlation, Understandability.
  \end{tabular} \qquad
  \begin{tabular}[b]{c}
    \includegraphics[width=.44\linewidth]{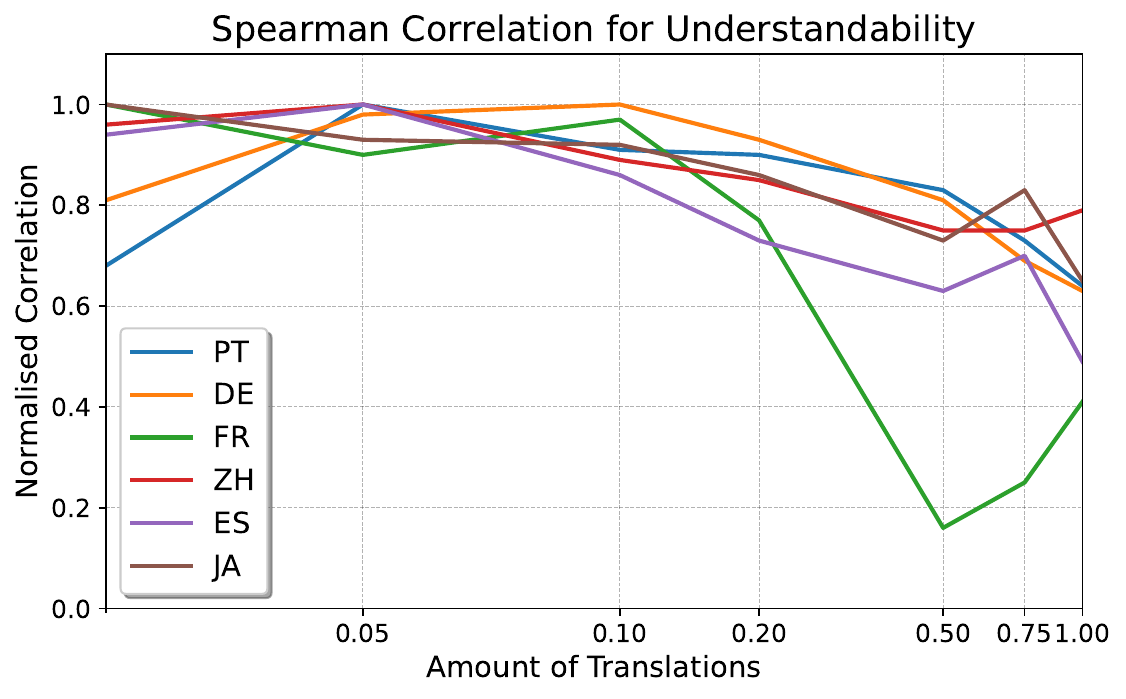} \\
    \small (b) Spearman Correlation, Understandability.
  \end{tabular}
    \\
  \begin{tabular}[b]{c}
    \includegraphics[width=.44\linewidth]{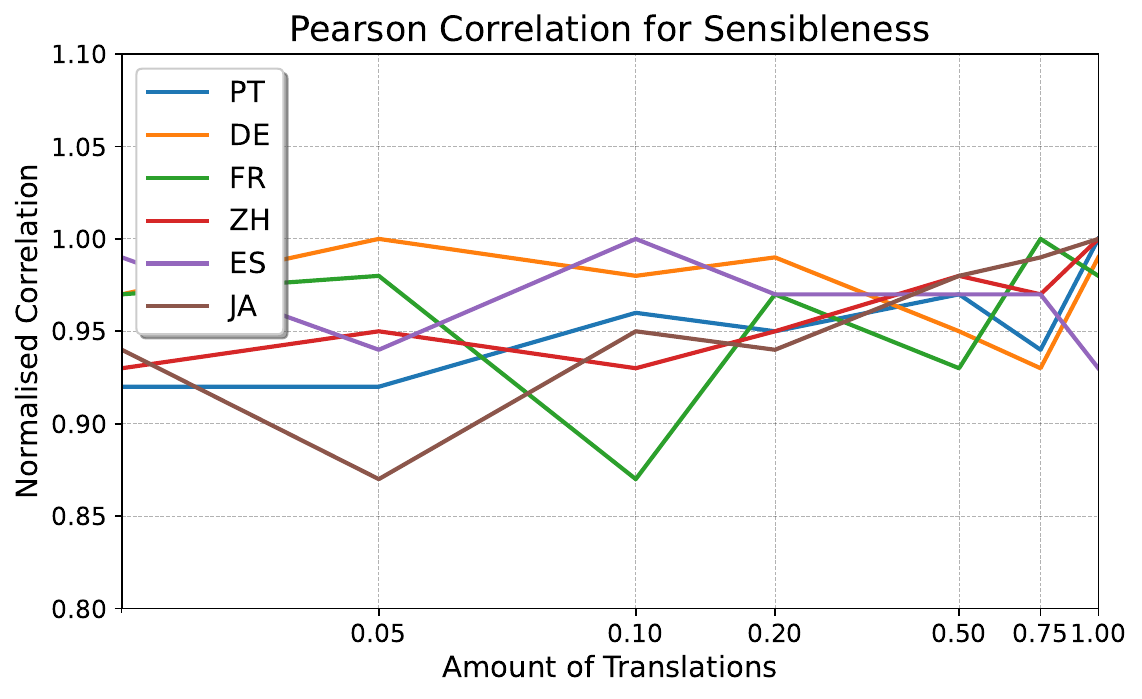} \\
    \small (c) Pearson Correlation, Sensibleness.
  \end{tabular} \qquad
  \begin{tabular}[b]{c}
    \includegraphics[width=.44\linewidth]{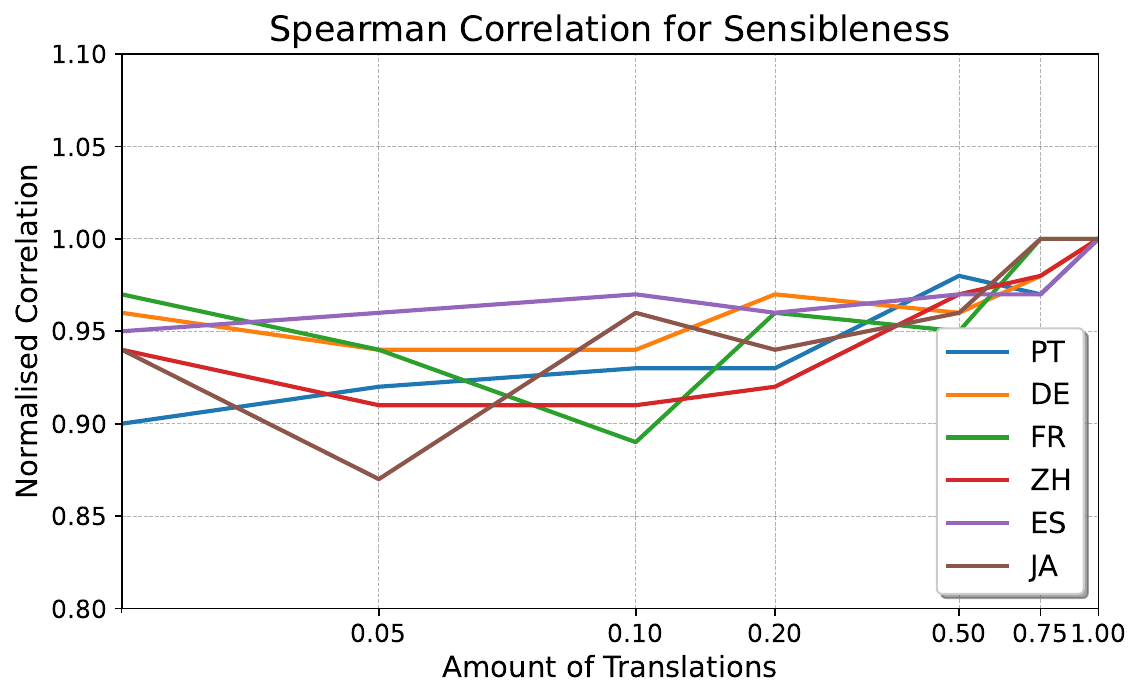} \\
    \small (d) Spearman Correlation, Sensibleness.
  \end{tabular}  
  \caption{Normalised Pearson and Spearman correlation for the Understandability and Sensibleness submetric with varying amount of translated training data. Numeric results available in Appendix \ref{sec:appendix_res}.}
  \label{fig:mtqe}
\end{figure*}

\section{MT Quality-aware finetuning}
\label{sec:qe_exp}

\begin{table}[ht]
\centering
\small
\begin{tabular}{l}
\hline
\textbf{EN:} Yes, I'd like to see the \underline{receipt}. \\ \hspace{0.6cm} Oh ! I see you \underline{bought} the \textcolor{blue}{watch} last week. \\
\textbf{PT:} Sim, gostava de ver o \textcolor{red}{receio}. \\ \hspace{0.6cm} Oh! Vejo\textcolor{red}{-te a fazer} o \textcolor{blue}{relógio} na semana passada.  \\
\textbf{QE score:} -0.670                                                                    \\ \hline
\hline
\textbf{EN:} Just \textcolor{blue}{look} around ? Ah, that's boring.                                             \\
\textbf{ES:} \textcolor{red}{¡¡¡¡¡¡¡¡¡¡¡¡¡¡¡¡¡¡¡¡¡¡¡¡¡¡¡¡¡¡¡¡}                                                    \\
\textbf{QE score:} -1.481                                                                    \\ \hline
\hline
\textbf{EN:} Eight \underline{tens}, six \underline{ones} and \underline{large silver} for others.                                   \\
\textbf{ZH:} \begin{CJK}{UTF8}{gbsn}八个十\textcolor{red}{个},六\textcolor{red}{个十个},其他\textcolor{red}{十个十个十个十个十}\end{CJK}...                                                 \\
\textbf{QE Score:} -1.312                                                                    \\ \hline
\end{tabular}
\caption{Examples of low quality translations with corresponding QE score. \textcolor{red}{Red} denotes MT error, with \underline{underline} in the source sentence indicating the closest alignment of the error. \textcolor{blue}{Blue} denotes keywords that refer to prior context.}
\label{tab:badmt}
\end{table}

\begin{table*}[ht]
\centering
\scriptsize
\begin{tabular}{lcccccccccccccccc}
\multicolumn{1}{l|}{}             & \multicolumn{2}{c}{\textbf{EN}}                    & \multicolumn{2}{c}{\textbf{PT}}                    & \multicolumn{2}{c}{\textbf{DE}}                    & \multicolumn{2}{c}{\textbf{FR}}                             & \multicolumn{2}{c}{\textbf{ZH}}                    & \multicolumn{2}{c}{\textbf{ES}}                    & \multicolumn{2}{c}{\textbf{JA}}                    & \multicolumn{2}{c}{\textbf{AVG}} \\
\multicolumn{1}{l|}{}             & \textbf{Pr.}  & \multicolumn{1}{c|}{\textbf{Sp.}}  & \textbf{Pr.}  & \multicolumn{1}{c|}{\textbf{Sp.}}  & \textbf{Pr.}  & \multicolumn{1}{c|}{\textbf{Sp.}}  & \textbf{Pr.}  & \multicolumn{1}{c|}{\textbf{Sp.}}           & \textbf{Pr.}  & \multicolumn{1}{c|}{\textbf{Sp.}}  & \textbf{Pr.}  & \multicolumn{1}{c|}{\textbf{Sp.}}  & \textbf{Pr.}  & \multicolumn{1}{c|}{\textbf{Sp.}}  & \textbf{Pr.}    & \textbf{Sp.}   \\ \hline
\multicolumn{17}{c}{\textbf{Understandability}}                                                                                                                                                                                                                                                                                                                                                                                                                  \\ \hline
\multicolumn{1}{l|}{\textbf{0 (EN)}}    & .376 & \multicolumn{1}{c|}{.187} & .366 & \multicolumn{1}{c|}{.167} & .328 & \multicolumn{1}{c|}{.172} & .351 & \multicolumn{1}{c|}{.120} & .318 & \multicolumn{1}{c|}{.202} & .342 & \multicolumn{1}{c|}{.204} & .204 & \multicolumn{1}{c|}{.176} & .327   & .194  \\
\multicolumn{1}{l|}{\textbf{5}}   & .403          & \multicolumn{1}{c|}{.182}          & .490          & \multicolumn{1}{c|}{.219}          & .344          & \multicolumn{1}{c|}{.172}          & \textbf{.385} & \multicolumn{1}{c|}{\textit{.091}} & .320          & \multicolumn{1}{c|}{\textbf{.235}} & \textbf{.429} & \multicolumn{1}{c|}{.236} & \textbf{.230} & \multicolumn{1}{c|}{.179} & \textbf{.372}   & .211  \\
\multicolumn{1}{l|}{\textbf{10}}  & .377          & \multicolumn{1}{c|}{.180}          & \textbf{.514} & \multicolumn{1}{c|}{.227}          & \textbf{.381} & \multicolumn{1}{c|}{.193}          & .294          & \multicolumn{1}{c|}{\textit{.091}}          & \textbf{.338} & \multicolumn{1}{c|}{.214}          & .385          & \multicolumn{1}{c|}{.212}          & .216          & \multicolumn{1}{c|}{.175}          & .358            & .206           \\
\multicolumn{1}{l|}{\textbf{20}}  & .384          & \multicolumn{1}{c|}{.177}          & .478          & \multicolumn{1}{c|}{.236}          & .333          & \multicolumn{1}{c|}{.203}          & .153          & \multicolumn{1}{c|}{\textit{.087}}          & .318          & \multicolumn{1}{c|}{.219}          & .315          & \multicolumn{1}{c|}{.214}          & .174          & \multicolumn{1}{c|}{.168}          & .308            & .202           \\
\multicolumn{1}{l|}{\textbf{50}}  & \textbf{.413} & \multicolumn{1}{c|}{.201} & .481          & \multicolumn{1}{c|}{\textbf{.242}} & \textbf{.381} & \multicolumn{1}{c|}{.213} & \textit{.103} & \multicolumn{1}{c|}{\textit{.053}}          & .310          & \multicolumn{1}{c|}{.200}          & .315          & \multicolumn{1}{c|}{.221}          & .219          & \multicolumn{1}{c|}{.149}          & .317            & .200           \\
\multicolumn{1}{l|}{\textbf{75}}  & .311          & \multicolumn{1}{c|}{.145}          & .247          & \multicolumn{1}{c|}{.211}          & .320          & \multicolumn{1}{c|}{.195}          & \textit{.047} & \multicolumn{1}{c|}{\textit{.048}}          & .163          & \multicolumn{1}{c|}{.149}          & .111          & \multicolumn{1}{c|}{.198}          & \textit{.108} & \multicolumn{1}{c|}{.127}          & .187            & .158           \\
\multicolumn{1}{l|}{\textbf{100}} & .336          & \multicolumn{1}{c|}{.117}          & .176          & \multicolumn{1}{c|}{.167}          & .262          & \multicolumn{1}{c|}{.150}          & \textit{.012} & \multicolumn{1}{c|}{\textit{.015}}          & .225          & \multicolumn{1}{c|}{.138}          & .117          & \multicolumn{1}{c|}{.158}          & \textit{.091} & \multicolumn{1}{c|}{\textit{.092}} & .174            & .126           \\ \hdashline
\multicolumn{1}{l|}{\textbf{ChatGPT}} & .397 & \multicolumn{1}{c|}{\textbf{.334}}          & .365          & \multicolumn{1}{c|}{.230}          & .332          & \multicolumn{1}{c|}{\textbf{.263}}          & .369          & \multicolumn{1}{c|}{\textbf{.273}}          & .276          & \multicolumn{1}{c|}{.182}          & .394 & \multicolumn{1}{c|}{\textbf{.263}}          & .228          & \multicolumn{1}{c|}{\textbf{.223}}          & .337            & \textbf{.263 }          \\ \hline
\multicolumn{17}{c}{\textbf{Sensibleness}}                                                                                                                                                                                                                                                                                                                                                                                                                       \\ \hline
\multicolumn{1}{l|}{\textbf{0 (EN)}}    & .658          & \multicolumn{1}{c|}{.676}          & .636          & \multicolumn{1}{c|}{.651}          & .657          & \multicolumn{1}{c|}{.655}          & .646 & \multicolumn{1}{c|}{.656}          & .640          & \multicolumn{1}{c|}{.656}          & \textbf{.646}          & \multicolumn{1}{c|}{.657}          & .590          & \multicolumn{1}{c|}{.599}          & .639            & .649           \\
\multicolumn{1}{l|}{\textbf{5}}   & .637          & \multicolumn{1}{c|}{.674}          & .629          & \multicolumn{1}{c|}{.632}          & .627          & \multicolumn{1}{c|}{.648}          & .637 & \multicolumn{1}{c|}{.656}                   & .629          & \multicolumn{1}{c|}{.646}          & .626          & \multicolumn{1}{c|}{.647}          & .567          & \multicolumn{1}{c|}{.596}          & .621            & .640           \\
\multicolumn{1}{l|}{\textbf{10}}  & .642          & \multicolumn{1}{c|}{.675}          & .639          & \multicolumn{1}{c|}{.664}          & .661          & \multicolumn{1}{c|}{.669}          & .636          & \multicolumn{1}{c|}{.661}                   & .637          & \multicolumn{1}{c|}{.656}          & .635          & \multicolumn{1}{c|}{.668}          & .575          & \multicolumn{1}{c|}{.604}          & .632            & .654           \\
\multicolumn{1}{l|}{\textbf{20}}  & .650          & \multicolumn{1}{c|}{.689}          & .627          & \multicolumn{1}{c|}{.670}          & .649          & \multicolumn{1}{c|}{.681}          & .627          & \multicolumn{1}{c|}{.666}                   & .621          & \multicolumn{1}{c|}{.661}          & .637          & \multicolumn{1}{c|}{.673}          & .568          & \multicolumn{1}{c|}{.614}          & .626            & .660           \\
\multicolumn{1}{l|}{\textbf{50}}  & .667          & \multicolumn{1}{c|}{.691}         & \textbf{.642} & \multicolumn{1}{c|}{.687}          & .650          & \multicolumn{1}{c|}{.672}          & .621          & \multicolumn{1}{c|}{.662}                   & .652          & \multicolumn{1}{c|}{.664}         & .629         & \multicolumn{1}{c|}{.673}         & .600 & \multicolumn{1}{c|}{\textbf{.642}} & .637            & .666           \\
\multicolumn{1}{l|}{\textbf{75}}  & .677 & \multicolumn{1}{c|}{.712} & .629          & \multicolumn{1}{c|}{\textbf{.694}} & .679 & \multicolumn{1}{c|}{\textbf{.702}} & .633          & \multicolumn{1}{c|}{\textbf{.679}}          & \textbf{.661} & \multicolumn{1}{c|}{\textbf{.673}} & .643 & \multicolumn{1}{c|}{\textbf{.695}} & .593          & \multicolumn{1}{c|}{.635}          & .645   & \textbf{.679}  \\
\multicolumn{1}{l|}{\textbf{100}} & .651          & \multicolumn{1}{c|}{.691}          & .606          & \multicolumn{1}{c|}{.675}          & .634          & \multicolumn{1}{c|}{.680}          & .605          & \multicolumn{1}{c|}{.669}                   & .642          & \multicolumn{1}{c|}{.667}          & .596          & \multicolumn{1}{c|}{.676}          & .599          & \multicolumn{1}{c|}{.637}          & .619            & .664           \\ \hdashline
\multicolumn{1}{l|}{\textbf{ChatGPT}} & \textbf{.746} & \multicolumn{1}{c|}{\textbf{.724}}          & .636          & \multicolumn{1}{c|}{.626}          & \textbf{.683}          & \multicolumn{1}{c|}{.675}          & \textbf{.695}          & \multicolumn{1}{c|}{.666}          & .655          & \multicolumn{1}{c|}{.645}          & .680 & \multicolumn{1}{c|}{.677}          & \textbf{.625}          & \multicolumn{1}{c|}{.610}          & \textbf{.674}            & .662\textbf{ }       \\ \hline
\end{tabular}
\caption{Average correlation results across 3 runs with different seeds for multilingual models when varying the amount of translated data.}
\label{tab:qe-res}
\end{table*}

The effects of noise introduced to the training data is a subject of intense research in the literature \cite{DBLP:conf/iclr/ZhangBHRV17, DBLP:conf/iclr/HuLY20, swayamdipta-etal-2020-dataset}. It is expected that, for this task, noise is introduced by low quality translations, reducing the performance of trained models. This issue was identified in Section \ref{sec:naive_exp}, where for the VSP model in particular, the models trained using translations performed much worse than the baseline approach. Our hypothesis is that some translations heavily disrupt morphosyntactic cues used to infer response fluency, as shown in Table \ref{tab:badmt}. We acknowledge that these low quality translations may also reduce the quality of the response by disrupting keywords that point to the context (which is important for Sensibleness), or even more subtle quality cues (e.g. loss of empathy, inconsistency with named entities). However, the NSP model is trained to discriminate between the original response and randomly selected response from the corpus. As such, the model's prediction will remain invariant to most translation errors.

These observations, paired with the fact encoder models only slightly underperform ChatGPT (a much larger and expensive model), motivate the work described in this section. We hypothesise that, by ameliorating the MT noise via identifying and filtering low quality translations, the encoder model performance can outperform LLMs such as ChatGPT, at a fraction of the cost.

Since there are no available references, an MT QE \citep{specia2018quality} automatic metric is used for this purpose. Formally, an MT QE model is a scoring function that assigns a score given a source sentence and hypothesis translation. The unboundedness and uncalibrated nature of this score across languages results in the need for a cumbersome analysis for each individual language in order to determine a threshold for filtering. Instead, we propose to use QE scores for response ranking, for each target language. This ensures a standardised method for filtering, improving the scalability of this method to new languages. 

\subsection{Experimental setup}  

\begin{figure}[ht]
  \centering
  \includegraphics[width=0.48\textwidth]{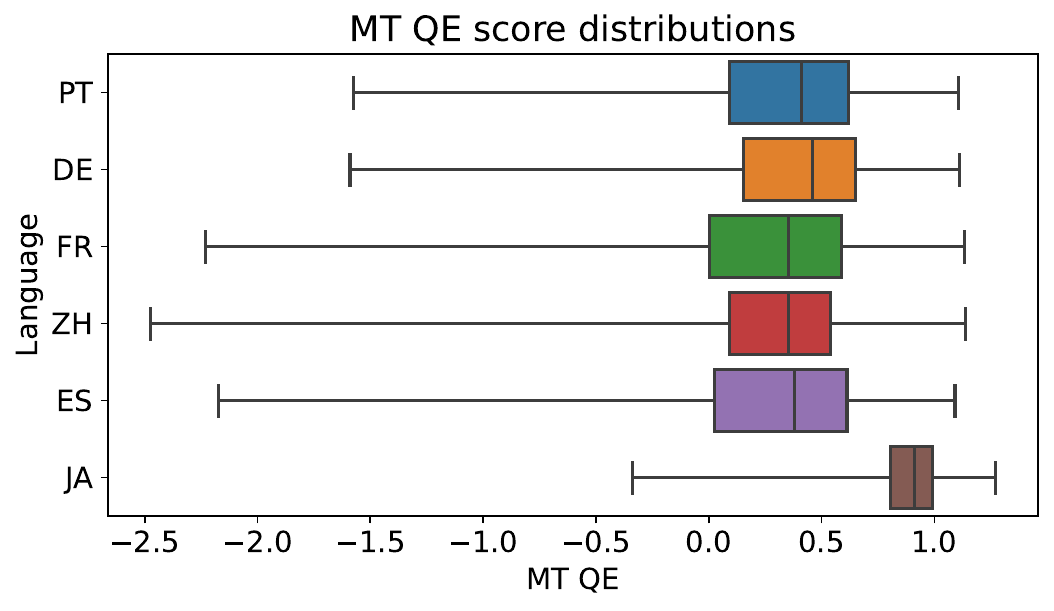}
  \caption{MT QE unnormalised score boxplot per language.}
  \label{fig:mtqedist}
\end{figure}

In order to confirm our hypothesis, we retrained all models using different amounts of translated data (100, 75, 50, 20, 10 and 5\%). The ranking of the translations was conducted by scoring them using the WMT20 COMET-QE-DA model \cite{rei-etal-2020-unbabels}. For the VSP model, we ranked the individual sentences, and then applied negative sampling. For the NSP model, we ranked the positive and negative samples separately and then merged them together. Figure \ref{fig:mtqedist} presents the unnormalised score boxplot per language for all sentences (context and responses) for DailyDialog.

One of the things we noticed when finetuning the monolingual models was that the VSP models had large variations in performance. This can be attributed to (1) the low amount of training data, especially when using very few examples (5\%, 10\%), and (2) low quality translations, which is the research question this experiment attempts to answer. Since the true impact of low quality translations is obfuscated by other factors, we decided to finetune the LANG models starting from the EN checkpoint instead of the pretrained XLM-RoBERTa, and include the zero-shot results as 0\%.

\subsection{Results}  

\paragraph{LANG}  
For the monolingual models, we plot normalised correlation results with the amount of MT data used during finetuning in Figure \ref{fig:mtqe}.
The \textbf{\textit{Understandability}} correlation results show that the optimal amount of translated data is language dependent, but with a clear indication that the inclusion of more translations decreases performance significantly. Instead, a lower amount of translations (5-10\%) yields optimal performance. This shows that this small finetuning step is essentially adapting a model that was already finetuned for the downstream task to the target-language domain.
For \textbf{\textit{Sensibleness}}, we see that the inclusion of more translations yields the best results. As such, we can conclude that low-quality MT does not adversely affect performance. We hypothesise this is due to MT being able to correctly translate keywords that indicate context awareness. Since we are only concerned about relevance, the overall sentence may still contain MT errors and be scored highly.

\paragraph{ML} The correlation results for the multilingual models are presented in Table \ref{tab:qe-res}.
For \textbf{\textit{Understandability}}, we note that, on average, and similar to LANG, the best performance is attained with the minimum amount of translated data (ML-5), with the performance decreasing when more translations are added. Comparing these results with ChatGPT, we observe an improvement in performance, but our encoder models are still generally weaker when using Spearman as a metric.
For \textbf{\textit{Sensibleness}}, decreasing the amount of data reduces the performance of the model. However, we note a decrease in performance when including the full amount of translated data (ML-100). This may be due to the inclusion of the worst translations -- typically hallucinations -- which is compounded by training on all languages. Unlike in Understandability, here we see that ChatGPT still outperforms the best encoder model in terms of Pearson correlation.

\subsection{Effect of low-quality translation during prediction}

\begin{figure*}[ht]
\centering
\subfloat[Understandability scatter plot.]{
  \centering
  \includegraphics[width=0.465\textwidth]{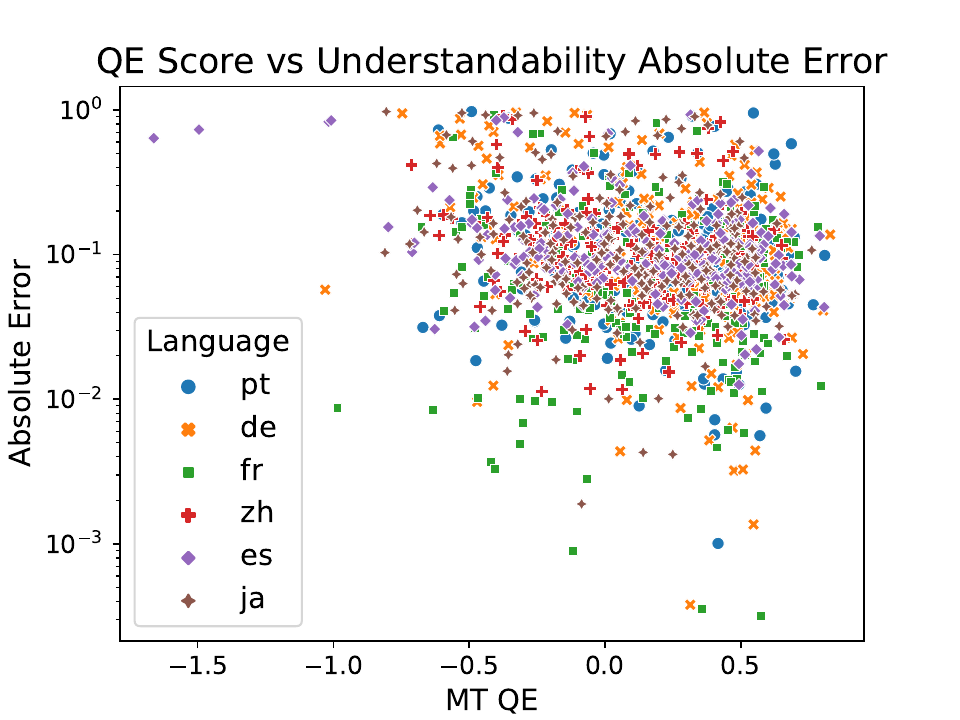}
  \label{fig:qevsU}}
\hspace{0.5cm}
\subfloat[Sensibleness scatter plot.]{
  \centering
  \includegraphics[width=0.465\textwidth]{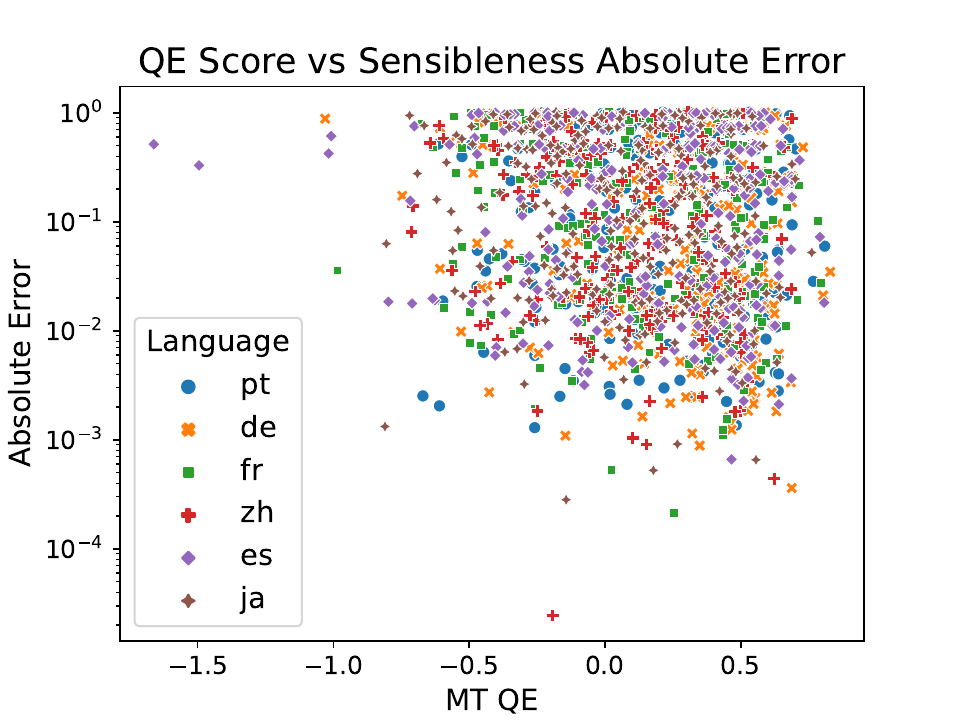}
  \label{fig:qevsS}}
\caption{Scatter plot comparing the test set MBART50 per-language QE z-scores (x-axis) versus the per sample Absolute Prediction Error (y-axis in log scale) for Understandability and Sensibleness subqualities.}
\label{fig:qevserror}
\end{figure*}

One might ask if a low-quality translation can induce the submetrics to output a different score. Intuitively, we hypothesise each model will attribute different scores in the face of low quality translations. More specifically, given the results presented in previous sections, we expect the test prediction error to be:

\begin{itemize}
    \item \textbf{Negatively correlated with the MT QE scores for VSP.} We know this model is highly sensitive to low quality translations, since MT errors frequently affect the fluency of the response (as identified in previous sections);
    \item \textbf{Weakly correlated for the NSP model.} The model showed robustness when including more translations during training, with performance decreasing only when we included all translations (ML-100) during training. 
\end{itemize}

In order to evaluate these assumptions, the correlation plots of the MT QE z-scores (obtained independently for each language) against the submetric absolute error using the best ML models (ML-5 for VSP and ML-75 for NSP) for the test set are presented in Figure \ref{fig:qevserror}.

For the \textbf{\textit{Understandability}} subquality, we note that there is a slight negative correlation between the absolute error and the MT QE score. This is also confirmed by a calculated Pearson Correlation value of -0.245. For the \textbf{\textit{Sensibleness}} subquality, the relationship between these two measures is less obvious. For instance, we note that, unlike for Understandability, maximum deviations are spread evenly across the QE scale, which points to the model erroneously predicting Sensibleness irrespective of the translation quality. Conversely, we also note a higher density of accurate predictions with lower QE scores. These results, paired with the calculated Pearson Correlation value of -0.129, confirm our hypothesis that the NSP model is more agnostic of MT quality than VSP.

\begin{table}[ht]
\centering
\small
\begin{tabular}{l}
\textbf{CTX:} Também me apercebi desta questão. E a\\automatização dos processos do escritório é essencial. \\
\textbf{RES:} Sim, fazer tudo manualmente demora demasiado.                                       \\\hdashline  
\textbf{EN-VSP:} .394 \hspace{1cm}\textbf{EN-NSP:} .824\\
\textbf{ML-VSP:} 1.00 \hspace{.95cm}\textbf{ML-NSP:} 1.00                                               \\
\textbf{Unders.:} 1.00 \hspace{1.1cm}\textbf{Sensibl:} 0.00                                                                       \\ \hline \hline

\textbf{CTX:} Ja, ich leite die Jungs am Kai.                                                              \\
\textbf{RES:} Wow, das klingt nach einem fantastischen Job, de\\du da bekommen hast.                       \\\hdashline  
\textbf{EN-VSP:} .963 \hspace{1cm}\textbf{EN-NSP:} .315\\\textbf{ML-VSP:} .941 \hspace{.95cm}\textbf{ML-NSP:} .981                                               \\
\textbf{Unders.:} 1.00 \hspace{1.1cm}\textbf{Sensibl:} 1.00

\end{tabular}
\caption{Examples of subquality predictions from the test set.}
\label{tab:testpred}
\end{table}

\subsection{Example test predictions}

We present representative examples of our best ML models' prediction (ML 5/75) in Table \ref{tab:testpred}. In the first example, the baseline English model fails to appropriately identify the understandability of the response. In the second example, we see that the multilingual model is able to correctly identify that the response takes into account the job presented in the context (manager) by complimenting it ("fantastic job"), which the EN model failed to identify. 

\section{Conclusions}
\label{sec:conclusions}

This paper explored the use of cross-lingual knowledge transfer for the novel task of automatic multilingual dialogue evaluation. We evaluated different strategies for this task, including zero-shot inference, MAD-X and Machine Translation augmentation. Empirically we showed that the naive approach of leveraging MT for augmentation is insufficient to outperform the baseline of English finetuning with a multilingual encoder-based LM, let alone a strong LLM. Instead, by filtering out low quality translations, we were able to reduce the gap of performance on ChatGPT, outperforming it on select correlation metrics. Experimental results showed that we obtain the best performance when training encoder models with the following proportions of MT-QE: 5\% for Understandability and 75\% for Sensibleness.

One could argue the notion of quality is intrinsically related to cultural norms. For instance, Japanese speakers may prefer a polite conversation, whereas German speakers might prefer a more direct interaction. A future research direction is to evaluate generative model responses in different languages using annotators exposed to the culture associated with a given language. In addition to ensuring the evaluation of the response meets the criteria of "quality" in different cultures, it would also allow for a qualitative analysis of the differences in the notion of quality between languages.

\section*{Limitations}

Perhaps the main limitation of this work is the restricted amount of languages studied. Ideally, we would have used a more comprehensible set of languages, including low-resource ones, to evaluate the consistency of the conclusions drawn from the experiments.

Another limitation is the focus on a single open-domain dialogue dataset. Dialogue evaluation metrics are known to correlate poorly when evaluated on unseen datasets \citep{yeh-etal-2021-comprehensive}. As such, it is not certain that the observations presented in this work would hold for other datasets, or even different annotations \citep{nsf}. 

Finally, the pretrained encoder, MT and QE models used in this work are not fully representative of all available models. We acknowledge that the optimal amount of filtering is likely to be different, depending on the combination of models used.

\section*{Ethics Statement}

This work leverages dialogues and annotations developed exclusively by English-speakers. This introduces an English-centric bias with respect to the notion of quality (and subqualities) in dialogues. Although not evaluated in depth in this work, there could be a chance that the models erroneously yield lower scores to responses not conforming to English notions of quality responses. 

The original dialogue dataset and generated responses were checked for personally identifiable information or offensive content by the original authors. Although highly unlikely, we acknowledge the translations may contain offensive content resulting from decoding.

The post-editing conducted in this work used a crowdsourcing platform that awarded users a fair wage according to their location.

\section*{Acknowledgements}

This research was supported by the Portuguese Recovery and Resilience Plan through project C645008882-00000055 (Responsible.AI), and by national funds through \textit{Fundação para a Ciência e a Tecnologia} (FCT) with references PRT/BD/152198/2021 and UIDB/50021/2020, and by the P2020 program MAIA (LISBOA-01-0247-FEDER-045909).

\bibliographystyle{acl_natbib}
\bibliography{custom,anthology}

\appendix

\begin{table*}[ht]
\centering
\scriptsize
\begin{tabular}{lcccccccccccccccc}

\multicolumn{1}{l|}{}             & \multicolumn{2}{c}{\textbf{EN}}                    & \multicolumn{2}{c}{\textbf{PT}}                    & \multicolumn{2}{c}{\textbf{DE}}                    & \multicolumn{2}{c}{\textbf{FR}}                             & \multicolumn{2}{c}{\textbf{ZH}}                    & \multicolumn{2}{c}{\textbf{ES}}                    & \multicolumn{2}{c}{\textbf{JA}}                    & \multicolumn{2}{c}{\textbf{AVG}} \\
\multicolumn{1}{l|}{}             & \textbf{Pr.}  & \multicolumn{1}{c|}{\textbf{Sp.}}  & \textbf{Pr.}  & \multicolumn{1}{c|}{\textbf{Sp.}}  & \textbf{Pr.}  & \multicolumn{1}{c|}{\textbf{Sp.}}  & \textbf{Pr.}  & \multicolumn{1}{c|}{\textbf{Sp.}}           & \textbf{Pr.}  & \multicolumn{1}{c|}{\textbf{Sp.}}  & \textbf{Pr.}  & \multicolumn{1}{c|}{\textbf{Sp.}}  & \textbf{Pr.}  & \multicolumn{1}{c|}{\textbf{Sp.}}  & \textbf{Pr.}    & \textbf{Sp.}   \\ \hline
\multicolumn{17}{c}{\textbf{Understandability}}                                                                                                                                                                                                                                                                                                                                                                                                                 \\ \hline
\multicolumn{1}{l|}{\textbf{0}}   & .347        & \multicolumn{1}{c|}{.192}        & .381          & \multicolumn{1}{c|}{.176}          & .353          & \multicolumn{1}{c|}{.184}          & \textbf{.349} & \multicolumn{1}{c|}{\textbf{.106}} & \textbf{.406} & \multicolumn{1}{c|}{.251}          & .372          & \multicolumn{1}{c|}{.210}          & .268          & \multicolumn{1}{c|}{\textbf{.223}} & .354           & .212          \\
\multicolumn{1}{l|}{\textbf{5}}   &              & \multicolumn{1}{c|}{}             & \textbf{.534} & \multicolumn{1}{c|}{\textbf{.259}} & .469          & \multicolumn{1}{c|}{.223}          & .347          & \multicolumn{1}{c|}{\textit{.095}} & .318          & \multicolumn{1}{c|}{\textbf{.263}} & \textbf{.459} & \multicolumn{1}{c|}{\textbf{.223}} & .236          & \multicolumn{1}{c|}{.208}          & \textbf{.387}  & \textbf{.231} \\
\multicolumn{1}{l|}{\textbf{10}}  &              & \multicolumn{1}{c|}{}             & .563          & \multicolumn{1}{c|}{.236}          & \textbf{.489} & \multicolumn{1}{c|}{\textbf{.227}} & .199          & \multicolumn{1}{c|}{\textit{.102}} & .300          & \multicolumn{1}{c|}{.233}          & .300          & \multicolumn{1}{c|}{.191}          & \textbf{.303} & \multicolumn{1}{c|}{.206}          & .357           & .218          \\
\multicolumn{1}{l|}{\textbf{20}}  &              & \multicolumn{1}{c|}{}             & .499          & \multicolumn{1}{c|}{.233}          & .356          & \multicolumn{1}{c|}{.211}          & .153          & \multicolumn{1}{c|}{\textit{.082}} & .323          & \multicolumn{1}{c|}{.223}          & .257          & \multicolumn{1}{c|}{.163}          & .251          & \multicolumn{1}{c|}{.191}          & .312           & .201          \\
\multicolumn{1}{l|}{\textbf{50}}  &              & \multicolumn{1}{c|}{}             & .433          & \multicolumn{1}{c|}{.214}          & .418          & \multicolumn{1}{c|}{.185}          & \textit{.117} & \multicolumn{1}{c|}{\textit{.017}} & .250          & \multicolumn{1}{c|}{.198}          & .233          & \multicolumn{1}{c|}{.140}          & .225          & \multicolumn{1}{c|}{.163}          & .289           & .175          \\
\multicolumn{1}{l|}{\textbf{75}}  &              & \multicolumn{1}{c|}{}             & .186          & \multicolumn{1}{c|}{.189}          & .306          & \multicolumn{1}{c|}{.158}          & \textit{.089} & \multicolumn{1}{c|}{\textit{.026}} & .319          & \multicolumn{1}{c|}{.198}          & .243          & \multicolumn{1}{c|}{.156}          & \textit{.226} & \multicolumn{1}{c|}{.185}          & .245           & .169          \\
\multicolumn{1}{l|}{\textbf{100}} &              & \multicolumn{1}{c|}{}             & .240          & \multicolumn{1}{c|}{.165}          & .347          & \multicolumn{1}{c|}{.144}          & \textit{.082} & \multicolumn{1}{c|}{\textit{.043}} & .248          & \multicolumn{1}{c|}{.206}          & .191          & \multicolumn{1}{c|}{.109}          & .216          & \multicolumn{1}{c|}{.146}          & .239           & .155          \\ \hline
\multicolumn{17}{c}{\textbf{Sensibleness}}                                                                                                                                                                                                                                                                                                                                                                                                                        \\ \hline
\multicolumn{1}{l|}{\textbf{0}}   & .621        & \multicolumn{1}{c|}{.654}        & .618          & \multicolumn{1}{c|}{.627}          & .667          & \multicolumn{1}{c|}{.668}          & .621          & \multicolumn{1}{c|}{.644}          & .605          & \multicolumn{1}{c|}{.647}          & .628          & \multicolumn{1}{c|}{.628}          & .577          & \multicolumn{1}{c|}{.592}          & .620           & .635          \\
\multicolumn{1}{l|}{\textbf{5}}   &              & \multicolumn{1}{c|}{}             & .615          & \multicolumn{1}{c|}{.636}          & .687          & \multicolumn{1}{c|}{.657}          & .632          & \multicolumn{1}{c|}{.628}          & .618          & \multicolumn{1}{c|}{.629}          & .599          & \multicolumn{1}{c|}{.631}          & .538          & \multicolumn{1}{c|}{.553}          & .616           & .626          \\
\multicolumn{1}{l|}{\textbf{10}}  &              & \multicolumn{1}{c|}{}             & .647          & \multicolumn{1}{c|}{.646}          & .672          & \multicolumn{1}{c|}{.655}          & .562          & \multicolumn{1}{c|}{.596}          & .607          & \multicolumn{1}{c|}{.626}          & \textbf{.635} & \multicolumn{1}{c|}{.637}          & .587          & \multicolumn{1}{c|}{.606}          & .619           & .630          \\
\multicolumn{1}{l|}{\textbf{20}}  &              & \multicolumn{1}{c|}{}             & .639          & \multicolumn{1}{c|}{.644}          & .680          & \multicolumn{1}{c|}{.679}          & .627          & \multicolumn{1}{c|}{.640}          & .620          & \multicolumn{1}{c|}{.633}          & .615          & \multicolumn{1}{c|}{.634}          & .582          & \multicolumn{1}{c|}{.595}          & .626           & .638          \\
\multicolumn{1}{l|}{\textbf{50}}  &              & \multicolumn{1}{c|}{}             & .651          & \multicolumn{1}{c|}{.680}          & .654          & \multicolumn{1}{c|}{.671}          & .601          & \multicolumn{1}{c|}{.631}          & .637          & \multicolumn{1}{c|}{.665}          & .613          & \multicolumn{1}{c|}{.639}          & .603          & \multicolumn{1}{c|}{.609}          & .626           & .647          \\
\multicolumn{1}{l|}{\textbf{75}}  &              & \multicolumn{1}{c|}{}             & .634          & \multicolumn{1}{c|}{.670}          & .640          & \multicolumn{1}{c|}{.681}          & \textbf{.643} & \multicolumn{1}{c|}{.664}          & .629          & \multicolumn{1}{c|}{.673}          & .615          & \multicolumn{1}{c|}{.639}          & .608          & \multicolumn{1}{c|}{.635}          & .627           & .656          \\
\multicolumn{1}{l|}{\textbf{100}} &              & \multicolumn{1}{c|}{}             & \textbf{.671} & \multicolumn{1}{c|}{\textbf{.693}} & \textbf{.681} & \multicolumn{1}{c|}{\textbf{.698}} & .631          & \multicolumn{1}{c|}{\textbf{.666}} & \textbf{.650} & \multicolumn{1}{c|}{\textbf{.688}} & .589          & \multicolumn{1}{c|}{\textbf{.659}} & \textbf{.617} & \multicolumn{1}{c|}{\textbf{.633}} & \textbf{.637}  & \textbf{.666} \\ \hline
\end{tabular}
\caption{Average correlation results for the monolingual models when varying the amount of translated data.}
\label{tab:qe-mono-res}
\end{table*}

\section{Training setup and Hyperparamters}
\label{sec:appendix_setup}

We used the XLM-R Large encoder model downloaded from HuggingFace \footnote{\url{huggingface.co/xlm-roberta-large}} for all experiments. A token representing the speaker was added for each turn, and a history length of 3 turns was used. We applied a regression head consisting of a 2-layer MLP with a hidden size of 1024 and a hyperbolic tangent function as activation for prediction. All parameters were trained/finetuned using Adam optimizer \cite{DBLP:journals/corr/KingmaB14}. 

The task adapters were trained using the recipe from \citet{mendonca-etal-2022-qualityadapt}, using a learning rate of 1e-4 and training for 10 epochs, with a batch size of 32. We used the existing language adapters from AdapterHub whenever possible (EN, ZH, JA) and trained the remaining using the AdapterHub's MLM recipe \footnote{\url{github.com/adapter-hub}} on Wikipedia data\footnote{\url{dumps.wikimedia.org}}. The fully finetuned models used a learning rate of 3e-6 and were trained for 3 epochs using a batch size of 16. Evaluation was conducted every 1,000 steps for the smaller training sets and 10,000 steps for the larger ones (75\% and 100 \%). The best performing model on the evaluation set was selected for testing. 

For the dialogue data preprocessing we used spaCy \footnote{\url{spacy.io}} and the corresponding core language models. For the translations we used \texttt{facebook/mbart-large-50-one-to-many-mmt} from HuggingFace. Batch size was set to 16 and decoding was conducted using beam search, with the number of beams set to 4.

We used a single Quadro RTX 6000 24GB GPU for all experiments.

\section{Additional Results}
\label{sec:appendix_res}

Table \ref{tab:qe-mono-res} presents the monolingual model results for the experiments of Section \ref{sec:qe_exp}. Due to time and computational constraints, we only conduct these experiments using a single seed.

\end{document}